\documentclass[11pt,a4paper]{article}
\usepackage[hyperref]{eacl2021}
\usepackage{times}
\usepackage{latexsym}

\usepackage{microtype}

\usepackage[T2A,T1]{fontenc}
\usepackage[utf8]{inputenc}
\usepackage[ukrainian,english]{babel}

\usepackage{float}
\usepackage{graphicx}

\aclfinalcopy 

\usepackage{multirow}

\title{UA-GEC: Grammatical Error Correction and Fluency Corpus for the Ukrainian Language\footnotemark}

\author{Oleksiy Syvokon \\
  Grammarly  \\
  \small\texttt{oleksiy.syvokon@gmail.com} \\\And
  Olena Nahorna \\
  Grammarly \\
  \small\texttt{olena.nahorna@grammarly.com} \\}

\date{}

\begin{document}
\maketitle
\begin{abstract}
We present a corpus professionally annotated for grammatical error correction (GEC) and fluency edits in the Ukrainian language. To the best of our knowledge, this is the first GEC corpus for the Ukrainian language. We collected texts with errors (20,715 sentences) from a diverse pool of contributors, including both native and non-native speakers. The data cover a wide variety of writing domains, from text chats and essays to formal writing. Professional proofreaders corrected and annotated the corpus for errors relating to fluency, grammar, punctuation, and spelling. This corpus can be used for developing and evaluating GEC systems in Ukrainian. More generally, it can be used for researching multilingual and low-resource NLP, morphologically rich languages, document-level GEC, and fluency correction. The corpus is publicly available at \url{https://github.com/grammarly/ua-gec}.
\end{abstract}

\footnotetext[1]{{This paper describes an older version of UA-GEC. A bigger  version of the corpus is available on GitHub. The paper update is in progress. }}

\begin{table*}[ht!]
    \centering
    \begin{tabular}{crrrrrc}
        \hline
        \textbf{Split} & \textbf{Writers} & \textbf{Texts} & \textbf{Sentences} & \textbf{Tokens} & \textbf{Annotations} & \textbf{Error rate} \\
        \hline
        Train   & 416 & 851 & 18,225 & 285,254 & 20,379 & 7.1\% \\
        Test    & 76  & 160 &  2,490 &  43,532 & 3,117 & 7.2\% \\
        \hline
        TOTAL   & 492 & 1,011 & 20,715 & 328,779 & 23,496 & 7.1\%\\
        \hline
    \end{tabular}
    \caption{The corpus statistics. Test split is independently annotated by two annotators (\emph{Error rate} is the average of the two in this case)}.
    \label{tab:splits}
\end{table*}

\section{Introduction}

Grammatical error correction (GEC) is a task of automatically detecting and correcting grammatical errors in written text. GEC is typically limited to making a minimal set of grammar, spelling, and punctuation edits so that the text becomes free of such errors. Fluency correction is an extension of GEC that allows for broader sentence rewrites to make a text more fluent---i.e., sounding natural to a native speaker \citep{sakaguchi-etal-2016-reassessing}.


Over the past decade, NLP researchers have been primarily focused on English GEC, where they indeed made substantial progress: F$_{0.5}$ score of the best performing model in the CoNLL-2014 shared task has increased from 37.33 in 2014 to 66.5 in 2020 \citep{ng-etal-2014-conll, omelianchuk-etal-2020-gector}. Multiple available datasets and shared tasks were a major contributing factor to that success: HOO-2011, HOO-2012, CoNLL-2013, CoNLL-2014, BEA-2019 \citep{dale-kilgarriff-2011-helping, dale-etal-2012-hoo, ng-etal-2013-conll, ng-etal-2014-conll, mizumoto-etal-2012-effect,napoles-etal-2017-jfleg,bryant-etal-2019-bea}.


However, languages other than English still present a set of challenges for current NLP methods. Mainstream models developed with English in mind are suboptimal for morphologically rich languages as well as languages with differing grammar \citep{tsarfaty-etal-2020-spmrl, ravfogel-etal-2018-lstm, hu2020xtreme, ahmad-etal-2019-difficulties}. The common issue is a scarcity of data---particularly high-quality annotated data that could be used for evaluation and fine-tuning.


More recently, the NLP community has started to pay more attention to non-English NLP \citep{ruder2020}. This positive recent trend manifests itself in the creation of new GEC corpora for mid- and low-resource languages: German, Czech, Russian, and Spanish, to name a few \citep{boyd-2018-using,naplava-straka-2019-grammatical,rozovskaya-roth-2019-grammar,davidson-etal-2020-developing}. These datasets are important to expand NLP research to new languages, explore the limitations of current models, and research new ways of training models in a low-resource setting.


Furthering that trend, we present a corpus annotated for grammatical errors and fluency in the Ukrainian language: UA-GEC. We first collected texts from a diverse pool of writers, both native and non-native speakers. The corpus covers a wide variety of domains: essays, social media posts, chats, formal writing, and more. We recruited professional proofreaders to correct errors related to grammar, spelling, punctuation, and fluency. We have made this corpus open source for the community\footnotemark[2] under the CC-BY 4.0 license.

\footnotetext[2]{\url{https://github.com/grammarly/ua-gec}}

To summarize, our contributions are as follows:

\begin{itemize}
 \item For the first time, diverse texts in Ukrainian are collected and annotated for grammatical, punctuation, spelling, and fluency errors.
 
 \item The corpus is released for public use under the CC-BY 4.0 licence.

\end{itemize}




\section{Data collection}

In this section, we describe the collection of texts with errors in the Ukrainian language. Section \ref{sec:annotation} will explain the annotation details.

\subsection{Statistics}

\begin{table}[ht]
    \centering
    \begin{tabular}{cccr}
        \hline
        \multicolumn{2}{c}{\textbf{Parameter}} & \textbf{Writers} & \textbf{Sent.} \\
        \hline
        \multirow{2}{4.4em}{Native}   & Yes     & 333   & 16,468 \\
                                      & No      & 163   & 4,247  \\ 
        \hline
        \multirow{3}{4.4em}{Gender}   & Female  & 288     & 9,601 \\
                                      & Male    & 205     & 11,044 \\
                                      & Other   & 5       & 70 \\
        \hline
        \multirow{3}{4.4em}{Background} & Technical     & 224 & 10,763 \\
                                        & Humanities    & 188 & 6,814 \\
                                        & Natural sci.  &  24 & 1,232 \\
                                        & Other         & 65  & 1,906 \\
        \hline
    \end{tabular}
    \caption{Profile of respondents}
    \label{tab:respondents}
\end{table}

We have collected 1,011 texts (20,715 sentences) written by 492 unique contributors. The average length of a text snippet is 20 sentences.

We partition the corpus into training and test sets. Each split consists of texts written by a randomly chosen disjoint set of people: all of a particular person's writing goes to exactly one of the splits. To better account for alternative corrections, we annotated the test set two times \citep{bryant-ng-2015-far}. The resulting statistics are shown in Table~\ref{tab:splits}.

In order to collect the data, we created an online form for text submission. All respondents who contributed to the data collection were volunteers. To attract a socially diverse pool of authors, we shared the form on social media. It contained a list of questions related to gender, native tongue, region of birth, and occupation, making it possible to further balance subcorpora and tailor them so they meet the purpose of various NLP tasks. Table \ref{tab:respondents} illustrates the profile of respondents based on some of these parameters.


\subsection{Collection tasks}

The online form offered a choice of three tasks: 1) writing an essay; 2) translating a fictional text fragment into Ukrainian; 3) submitting a personal text. Our goal was to collect a corpus of texts that would reflect errors typically made by native and non-native speakers of Ukrainian. Therefore, before performing a task, the respondents were asked not to proofread their texts as well as to refrain from making intentional errors. Each task varied in the number of requirements.

\begin{table}[h]
    \centering
    \small
    \begin{tabular}{p{22em}}
        \hline
        Write an essay on the topic "What's your favorite animal?" Genre: fictional. In the essay, state: what your favorite animal is; what it looks like; why you like this particular animal; whether you would like to keep it at home. Volume: about 15 sentences. \\  
        
        \hline
        
        Write a letter of complaint. Recipient: a restaurant administrator. Genre: formal. In the letter, state: the date of your visit to the restaurant; the reason for your complaint; your suggestions about how the restaurant could improve its service. Volume: about 15 sentences. \\
        \hline
    \end{tabular}
    \caption{Examples of the essay prompts. In total, there were 20 prompts in the \emph{Essay} task.}
    \label{tab:essay_prompts}
\end{table}

\textbf{Essays.} Respondents were offered one of twenty essay topics, each stipulating the genre, length, and structure of the essay. We chose from among the most common topics for essays (e.g., “What was your childhood dream?” or “Write a birthday party invitation”) not requiring a profound knowledge of a certain subject, which made it easy for the respondents to produce texts. Each essay was supposed to be written in accordance with one of four genres: formal, informal, fictional, or journalistic. The scientific genre was excluded as a potential writing blocker due to its inherent complexity: such texts require an author’s preliminary preparation and therefore cannot be written on an ad hoc basis. Specification of the genre allowed us to moderate the heterogeneity of the corpus. Besides topic and genre requirements, each task description contained prompts---i.e., prearranged points to cover in the text that facilitated text production. Refer Table~\ref{tab:essay_prompts} for essay prompts examples.

\textbf{Translation of fictional texts.} Fictional text fragments were taken from public domain books written by classic authors in five languages: English, French, German, Polish, and Russian. The rationale behind suggesting translation from a range of foreign languages was to diversify the errors made by respondents as a result of L1 interference.

\textbf{Personal texts.} Unlike the aforementioned tasks, personal text submission was not explicitly regulated: respondents could submit texts of any genre, length, or structure. However, no more than 300 sentences submitted by a unique person were added to the corpus. This was done to balance the corpus from an idiolect perspective.

UA-GEC is mostly composed of personal texts (77\%); fictional texts translations rank second (19\%), and essays are the least numerous (4\%).

\section{Data annotation}\label{sec:annotation}

The Ukrainian language is a fusional language. Notional parts of speech possess a rich inflectional paradigm, which is why Ukrainian admits free word order.

We enrolled two annotators on the project, both native speakers of Ukrainian with a degree in Ukrainian linguistics. One of them is a freelance editor, the other a teacher of Ukrainian.

The annotation process was split into two subtasks: first, annotators performed error identification and correction; after that, error labeling. We found that doing these subtasks sequentially was faster than performing them in a combined mode. The same annotators performed both tasks.

\subsection{Annotation format}

The categorized errors in the processed data are marked by the following in-text notations:
{\small\verb|{error=>edit:::Tag}|}, where {\small\verb|error|} and {\small\verb|edit|} stand for the text item before and after correction, respectively, and {\small\verb|Tag|} denotes an error category (Grammar, Spelling, Punctuation, or Fluency). Table~\ref{tab:annotation_examples} lists example sentences annotated for each category.

\begin{table*}[t!]
    \centering
    \setlength{\tabcolsep}{18pt}
    
    \begin{tabular}{lc}
        \hline
        \textbf{Error type} & \textbf{Example} \\
        \hline
        \multirow{2}{6em}{Grammar} &
        \begin{otherlanguage*}{ukrainian}
        {\small\verb|Я {йти=>ходжу:::Grammar} до школи.|}
        \end{otherlanguage*} \\
        & {\small\verb|I {goes=>go:::Grammar} to school.|} \\
        \hline

        \multirow{2}{6em}{Spelling} &
        \begin{otherlanguage*}{ukrainian}
        {\small\verb|Він {хотв=>хотів:::Spelling} поговорити.|}
        \end{otherlanguage*} \\
        & {\small\verb|He {wnted=>wanted:::Spelling} to talk.|} \\
        \hline
        
        \multirow{2}{6em}{Punctuation} &
        \begin{otherlanguage*}{ukrainian}
        {\small\verb|Ти будеш завтра вдома {=>?:::Punctuation}|}
        \end{otherlanguage*} \\
        & {\small\verb|Are you going to be home tomorrow {=>?:::Punctuation}|} \\
        \hline
        
        \multirow{2}{6em}{Fluency} &
        \begin{otherlanguage*}{ukrainian}
        {\small\verb|{Існуючі =>Теперішні:::Fluency} ціни дуже високі.|}
        \end{otherlanguage*} \\
        & {\small\verb|{Existing=>Current:::Fluency} prices are very high.|} \\

    \end{tabular}
    \caption{Examples of annotation in each error category}
    \label{tab:annotation_examples}
\end{table*}

Besides error correction and labeling, the annotators were asked to identify sensitive content---i.e., sentences containing pejorative lexis or perpetuating bias related to race, gender, age, etc. Such sentences are marked in the metadata, which enables simple data filtering to debias it by the stated criteria.
The GitHub repository contains a detailed description of the annotation scheme along with a Python library to process the corpora.

\subsection{Error categories}

\newcommand{\errortype}[1]{{\small{\texttt{#1}}}}


The annotators labeled corrections with one of the following high-level error categories: Grammar, Spelling, Punctuation, and Fluency. Table~\ref{error-categories} demonstrates the error distribution by category. 

\errortype{Spelling} accounts for 20.8\% of all corrections. This is similar to RULEC-GEC \citep{rozovskaya-roth-2019-grammar}, where the portion of spelling errors is 21.7\%. \errortype{Punctuation} edits (39.6\%) are more frequent than in other corpora (for example, in the W\&I corpus \citep{bryant-etal-2019-bea}, \errortype{Punctuation} is 17\%). We explain this by the fact that in the Ukrainian language, punctuation rules are sharply defined; thus, a lot of punctuation marks are frequently misused, especially commas. Also, there were a large number of typographical fixes, like replacing a dash (``-'') with an em-dash (``---'') where appropriate. \errortype{Grammar} accounts for 15.6\% of all errors.

\begin{table}[ht]
\centering
\begin{tabular}{lrcc}
\hline \textbf{Error type} & \textbf{Total}  & \textbf{\%} & \textbf{\raisebox{4pt}[13pt][8pt]{\vtop{\hbox{Per 1000}\hbox{tokens}}}} \\ \hline
Fluency & 5,561  & 20.8 & 16.9 \\
Grammar & 3,726 & 15.6 & 11.3 \\
Punctuation & 9,314 &  & 28.3 \\
Spelling & 4,895 & 20.8 & 14.8 \\
\hline
TOTAL & 23,496 & 100.0 & 71.5 \\
\hline
\end{tabular}
\caption{\label{error-categories} Error distribution by category }
\end{table}

\textbf{Fluency}.The fluency category embraces error types that have to do with the inaccurate use of lexical or structural units. The former has to do with the correction of violated collocational preference, contaminated idioms, unconventional pleonasms and tautologies, calques, etc. The latter concerns rewriting lengthy syntactic structures or syntactic structures whose word order hinders comprehension — even though such structures do not break grammatical rules, style guides recommend they should be avoided in order to improve readability.

Close analogs of this category presented in related datasets are as follows: \errortype{Lexical choice} in RULEC-GEC \citep{rozovskaya-roth-2019-grammar}; \errortype{lex} (error in lexicon or phraseology), \errortype{stylColl} (colloquial expression), \errortype{stylOther} (bookish, dialectal, slang, hyper-correct expression), \errortype{stylMark} (redundant discourse marker), \errortype{fwNC} (foreign word) in AKCES-GEC \citep{naplava-straka-2019-grammatical}; \errortype{Vocabulary} and \errortype{Coherence/Cohesion} in MERLIN \citep{boyd-2018-using}. 

\errortype{Fluency} accounts for 24.5\% of all errors. This may be attributed to the fact that around 30\% of respondents were not native Ukrainian speakers and therefore used a lot of calques, both lexical and structural, from other languages. Another reason is style correction: annotators corrected non-standard language into standard one to make the text sound more fluent and natural.

\subsection{Inter-annotator agreement}

\begin{table}[H]
    \centering
    \begin{tabular}{cccc}
        \hline
        \textbf{Pass 1} & \textbf{Pass 2} & \textbf{Error rate} & \textbf{Unchanged} \\
        \hline
        Ann. A & Ann. B & 2.9\% & 64\% \\
        Ann. B & Ann. A & 1.2\% & 75\% \\
        \hline
    \end{tabular}
    \caption{Inter-annotator agreement based on the second-pass proofreading. \textit{Error rate} is the density of annotations made on the already corrected text. \textit{Unchanged} is the percentage of sentences that have not been changed on the second pass. }
    \label{tab:agreement_2}
\end{table}



We follow the \citet{rozovskaya-roth-2010-annotating} setup for computing the inter-annotator agreement. A text that was corrected by one annotator is passed to the other annotator. \textit{Agreement} then is the percentage of sentences that did not require any changes during the second pass. This metric is important, given that our goal is to make a sentence well-formed, no matter whether the annotators propose the same changes \citep{rozovskaya-roth-2019-grammar}. We run this evaluation on a set of 200 sentences. Table \ref{tab:agreement_2} shows that 64\% of sentences corrected by Annotator A remained unchanged after the Annotator B's pass. The error rate has dropped from 7.1\% to 2.9\% errors. Similarly, Annotator A that proofreads after Annotator B leaves 75\% of sentences unchanged.

This inter-annotator agreement (64\%/75\% of unchanged sentences) may seem low, but it is in line with other GEC corpora: for English the reported numbers are 37\%/59\%, for Russian they are 69\%/91\% \citep{rozovskaya-roth-2010-annotating,rozovskaya-roth-2019-grammar}.


\subsection{Comparison to other GEC datasets}

Table~\ref{tab:related_corpora} lists statistics of our corpus in relation to similar GEC corpora in other languages. We refer the reader to \citep{naplava-straka-2019-grammatical} and \citep{rozovskaya-roth-2019-grammar} for good GEC corpora surveys.

\begin{table}[htpb]
    \centering
    \resizebox{7.5cm}{!}{%
    \begin{tabular}{lcrr}
        \hline \textbf{Language} & \textbf{Corpus} & \textbf{Sent.} & \textbf{Er.} \\
        \hline
        \multirow{4}{3em}{English} & Lang-8 & 1,147,451 & 14.1 \\
            & NUCLE  & 57,151 & 6.6 \\
            & FCE    & 33,236 & 11.5 \\
            & W\&I+L & 43,169 & 11.8 \\
            & JFLEG  & 1,511  & \\
            & CWEB   & 13,574 & 1.74 \\
        Czech   & AKCES-GEC & 47,371  & 21.4 \\
        German  & Falko-MERLIN & 24,077 & 16.8 \\
        Romanian & RONACC & 10,119 &  \\
        Russian & RULEC-GEC & 12,480 & 6.4 \\
        Spanish & COWS-L2H \footnotemark[3] & 12,336  & \\
        \hline
        \textbf{Ukrainian} & \textbf{UA-GEC} & \textbf{20,715} & \textbf{7.1} \\
        \hline
        
    \end{tabular}
    }
    \caption{Statistics of related GEC corpora. \emph{Er.} is the error rate, in percent. This work is highlighted in bold.}
    \label{tab:related_corpora}
\end{table}

\footnotetext[3]{{At the time of writing, COWS-L2H was under construction. The statistics is for March 2021}}

\section{Conclusion}

Having high-quality corpora is of paramount importance for advancing the state of multilingual NLP. The novelty of our work consists in releasing the first professionally annotated corpus, which will facilitate further development of grammatical error correction and fluency edits in the Ukrainian language. The corpus is made publicly available at \url{https://github.com/grammarly/ua-gec} under the CC-BY 4.0 license.

%

\section*{Acknowledgments}

This work is supported by Grammarly. We thank Nastia Osidach, Ira Kotkalova, Anna Vesnii, Halyna Kolodkevych, and everyone else who participated in the corpus creation.

\bibliography{eacl2021}
\bibliographystyle{acl_natbib}

\end{document}